\documentclass{article}

\usepackage{microtype}
\usepackage{graphicx}
\usepackage{subfigure}
\usepackage{booktabs} 

\usepackage{hyperref}

\usepackage[accepted]{icml2024}

\usepackage{amsmath}
\usepackage{amssymb}
\usepackage{mathtools}
\usepackage{amsthm}
\usepackage{caption}
\usepackage{bbm}
\usepackage{subcaption,graphicx}
\usepackage{subfigure}
\usepackage{pifont}
\usepackage{multirow}
\usepackage{booktabs}
\usepackage{array}
\DeclareMathOperator*{\argmin}{arg\,min}

\usepackage[capitalize,noabbrev]{cleveref}

\theoremstyle{plain}

\theoremstyle{definition}

\theoremstyle{remark}

\usepackage[textsize=tiny]{todonotes}

\icmltitlerunning{Post-processing fairness with minimal changes}

\begin{document}

\twocolumn[
\icmltitle{Post-processing fairness with minimal changes}



\icmlsetsymbol{equal}{*}

\begin{icmlauthorlist}
\icmlauthor{Federico Di Gennaro}{equal,axa}
\icmlauthor{Thibault Laugel}{equal,axa,lip6}
\icmlauthor{Vincent Grari}{axa,lip6}
\icmlauthor{Xavier Renard}{axa,lip6}
\icmlauthor{Marcin Detyniecki}{axa,lip6,polish}
\end{icmlauthorlist}

\icmlaffiliation{axa}{AXA, Paris, France}
\icmlaffiliation{lip6}{TRAIL, Sorbonne Université, Paris, France}
\icmlaffiliation{polish}{Polish Academy of Science, IBS PAN, Warsaw, Poland}

\icmlcorrespondingauthor{Thibault Laugel}{thibault.laugel@axa.com}

\icmlkeywords{Machine Learning, ICML}

\vskip 0.3in
]



\printAffiliationsAndNotice{\icmlEqualContribution} 

\begin{abstract}
In this paper, we introduce a novel post-processing algorithm that is both model-agnostic and does not require the sensitive attribute at test time. In addition, our algorithm is explicitly designed to enforce minimal changes between biased and debiased predictions—a property that, while highly desirable, is rarely prioritized as an explicit objective in fairness literature. 
Our approach leverages a multiplicative factor applied to the logit value of probability scores produced by a black-box classifier. We demonstrate the efficacy of our method through empirical evaluations, comparing its performance against other four debiasing algorithms on two widely used datasets in fairness research.
\end{abstract}
\vspace{-0.5cm}

\section{Introduction and Context}
\label{sec:introduction}

The increasing adoption of machine learning models in high-stake applications - e.g health care \cite{de2018clinically}, criminal justice \cite{kleinberg2016inherent} or credit lending \cite{bruckner2018promise} - has led to a rise in the concern about the potential bias that models reproduce and amplify against historically discriminated groups. 
Addressing these challenges, the field of algorithmic fairness has seen numerous bias mitigation approaches being proposed, tackling the problem of unfairness through drastically different angles~\cite{Romei2013AMS}.
Among them, \textit{post-processing} bias mitigation techniques (see Table~\ref{tab:comparison} for an overview) focus on debiasing the predictions of a trained classifier, rather than training a new, fair, model from scratch. Generally relying on label flipping heuristics~\cite{ROC,hardt2016equality} or optimal transport theory~\cite{jiang2020wasserstein}, they have been argued to achieve comparable performance~\cite{cruz2023unprocessing}
while generally requiring less computation and performing fewer modifications to the original predictions than their pre-processing and in-processing counterparts~\cite{jiang2020wasserstein,krco2023mitigating}. 

\newcolumntype{C}[1]{>{\centering\arraybackslash}p{#1}}
\begin{table*}[!t]
  \centering
    \begin{tabular}{l|C{2cm}C{2cm}C{2cm}C{2cm}}
    \toprule
    \bfseries \parbox{2cm}{Paper} & \bfseries \parbox{2cm}{\centering Model \\ agnostic} & \bfseries \parbox{2cm}{\centering No Sensitive \\ at test time} & \bfseries \parbox{2cm}{\centering Metric \\ optimized} & \bfseries \parbox{2cm}{\centering Minimizes \\ changes} \\
    \midrule
    \midrule
    ROC \cite{ROC} & \ding{51} & \ding{55} & DP & \ding{55} \\
    EO post-processing \cite{hardt2016equality} & \ding{51} & \ding{55} & EO & \ding{55} \\
    IGD post-processing \cite{lohia2019bias} & \ding{51} & \ding{55} & DP & \ding{55} \\
    Wass-1 post-processing \cite{jiang2020wasserstein} & \ding{51} & \ding{55} & DP & \ding{51}\\
    Wass-1 Penalized LogReg \cite{jiang2020wasserstein} & \ding{55} (LogReg) & \ding{51} & DP & \ding{51}\\
    FST \cite{wei20a} & \ding{51} & \ding{51} & DP, \ EO & \ding{55} \\
    RNF \cite{du2021fairness} & \ding{55} (NNs) & \ding{51} & DP, \ EO & \ding{55} \\
    FCGP \cite{nguyen2021fairness} & \ding{51} & \ding{55} & DP & \ding{55} \\
    FairProjection \cite{alghamdi2022beyond} & \ding{51} & \ding{55} & DP, \ EO & \ding{55}\\
    LPP (sensitive-unaware) \cite{xian2024optimal} & \ding{51} & \ding{51} & DP, \ EO & \ding{55}\\
    LPP (sensitive-aware) \cite{xian2024optimal} & \ding{51} & \ding{55} & DP, \ EO & \ding{55} \\
    \bottomrule
    \textbf{RBMD (Ours)} & \ding{51} & \ding{51} & DP, \ EO & \ding{51}\\
    \bottomrule
    \end{tabular}
  \caption{Comparison between post-processing methods.}
  \label{tab:comparison}
  \vspace{-0.4cm}
\end{table*}

Although rarely discussed, expecting the debiasing method to perform a low number of prediction changes is especially interesting in contexts where fairness is enforced while a model is already in production~\cite{krco2023mitigating}. In real-world applications, maintaining the integrity and reliability of predictive models is crucial, especially when they have undergone rigorous validation and expert review. For example, in non-life insurance pricing, experts 
commonly employ Generalized Additive Models (GAMs) with splines or polynomial regression on Generalized Linear Models to ensure that price 
are justifiable and align with both business objectives and customer expectations (e.g., avoiding price increases that could negatively impact customer satisfaction and brand reputation). 
Ensuring debiasing methods do not significantly alter validated predictions can therefore be essential. Yet, surprisingly, this property is rarely introduced explicitly in the objectives of existing post-processing algorithms. Instead, it is generally implicitly integrated: e.g., methods relying on label-flipping~\cite{ROC} aim to target only the "required" instances to enhance fairness. 
Additionally, the vast majority of existing approaches generally suffer from two major drawbacks, as shown in Table~\ref{tab:comparison}. First, most approaches require the sensitive attribute at inference for fairness (e.g., \citet{hardt2016equality,lohia2019bias,nguyen2021fairness}). However, this is unrealistic in many practical settings where the sensitive attribute is unavailable at test time. Second, among the works that avoid this assumption, one~\cite{jiang2020wasserstein} is limited to linear settings and \citet{du2021fairness} is not model agnostic. Concurrently to our work, \citet{tifreafrappe} proposed FRAPPE, recognizing and addressing the same limitations. 

In this short paper, we propose to address these issues by introducing a new post-processing method that is: (i) \textbf{model-agnostic}, (ii) \textbf{does not require the sensitive attribute at test time}, and (iii) \textbf{explicitly minimizes the number of prediction changes}.
To do so, we frame the post-processing task as a new supervised learning problem taking as input the previously trained (biased) model. We address this problem by introducing a new approach, RBMD (\emph{Ratio-Based Model Debiasing}), which predicts a multiplicative factor to rescale the biased model's predictions such that they better satisfy fairness guarantees.  
This allows us to leverage techniques traditionally used by in-processing methods to mitigate bias, such as the adversarial reconstruction of the sensitive attribute~\cite{zhang2018mitigating}, which does not require having the sensitive attribute during the inference. Experiments show that this allows RBMD to achieve competitive results in terms of fairness and accuracy scores compared to existing approaches while altering fewer predictions.

\section{Proposition}
\label{sec:ratio}
\subsection{Post-processing as a supervised learning task} 
\label{subsec:problem statement}

Let $\mathcal{X}$ be an input space consisting of $d$ features and $\mathcal{Y}$ an output space. Let us consider a traditional algorithmic fairness framework in which we want to predict an output variable $Y\subset\mathcal{Y}$ given an input variable $X\subset\mathcal{X}$ while being unbiased from a sensitive variable $S$. For the sake of simplicity, we decide in this paper to focus on a binary classification problem where $\mathcal{Y}=\{0,1\}$ and for which also the sensitive attribute $S$ takes values on a binary sensitive space $\mathcal{S}$. 
Finally, we suppose to have access to a black-box model $f \colon \mathcal{X} \rightarrow [0,1]$ trained on a training set $\mathcal{D}_{\text{train}}=\{(s_i, x_i, y_i)\}_{i=1}^N$ that consists of $N$ realizations of the triplet $(S,X,Y)$. The black-box model $f$ is trained for a binary classification task and outputs the probability of belonging to the class $Y=1$. We finally consider the prediction coming from the scores outputted by $f$ as a random variable $\hat{Y}^f=\mathbbm{1}_{f(X)>0.5}$. \\
Given a score $f(X)\in [0,1]$ or a prediction $\hat{Y}^f$, the goal of a post-processing task is to learn a new predictions $\hat{Y}$ that is more fair with respect to some fairness notion. In this paper, we focus on \emph{Demographic Parity}\footnote{We describe in Section~\ref{subsec:proposal} how to adapt our approach to \emph{Equalized Odds}. However, implementing and testing this is beyond the scope of this short paper.}, measured using the \emph{P-rule} criterion: \vspace{-0.1cm}
\begin{equation}
    \textit{P-Rule} = \min \left( \frac{\mathbb{P}(\hat{Y}=1\mid S=1)}{\mathbb{P}(\hat{Y}=1\mid S=0)}, \ \frac{\mathbb{P}(\hat{Y}=1\mid S=0)}{\mathbb{P}(\hat{Y}=1\mid S=1)} \right).
    \nonumber
    \vspace{-0.1cm}
\end{equation}
Most works address this problem by learning a new mapping $\hat{Y}^f \rightarrow \hat{Y}$ at inference time. Although this saves computation time by not requiring learning a new model like in-processing methods, it generally relies on solving an optimization problem for each new batch of predictions (or scores) $\hat{Y}^f$, and usually imposes the availability of the sensitive attribute at test time (cf. Table~\ref{tab:comparison}). Furthermore, as the true labels $Y$ are generally unavailable at inference time, no constraint on the accuracy of the new predictions $\hat{Y}$ can be imposed: the relationship between $\hat{Y}$ and $Y$ is then only indirectly preserved by the fact that minimal changes are performed; an objective which is also not consistently imposed (cf. Table~\ref{tab:comparison}). For these reasons, rather than learning a new mapping for each vector $\hat{Y}^f$ at inference time, we propose to train a model $g:[0,1]\times \mathcal{X} \rightarrow\mathcal{Y}$, taking as input both $X$ and its score $f(X)\in[0,1]$ to automatically perform these changes. In the rest of the paper, we refer to $g(X,f(X))$ as $g(X)$ and we define the prediction coming from the score $g(X)$ as $\hat{Y}^g=\mathbbm{1}_{g(X)>0.5}$.
Following the motivations described in Section~\ref{sec:introduction}, the model~$g$ should therefore satisfy the following desiderata: the model should be [1] \textbf{accurate}, [2] \textbf{fair} and [3] \textbf{perform as few changes as possible} to the predictions of $f$. 
The resulting optimization problem we address in this paper can thus be written under Demographic Parity and for $\epsilon$ and $\eta$ strictly positive as:
\vspace{-0.1cm}
\begin{equation}
    \begin{aligned}
        \min_{g}  \quad \mathbbm{E} [\mathcal{L}_Y(\hat{Y}^g, Y)] \quad \textrm{[1]}\\
        \textrm{s.t} \quad |\mathbbm{E}[\hat{Y}^g | S=1] - \mathbbm{E}[\hat{Y}^g | S=0] | \leq \epsilon \quad \textrm{[2]}\\
        \textrm{and} \quad \mathbbm{E}[ \mathbbm{1}_{\hat{Y}^g\neq \hat{Y}^f}] \leq \eta \quad \textrm{[3]}
    \end{aligned}
    \vspace{-0.1cm}
    \label{eq:optimization-pb}
\end{equation} 
with $\mathcal{L}_Y$ the binary cross-entropy, and $\epsilon$ and $\eta$ controlling the level of unfairness and changes allowed. 

\subsection{Ratio-Based Model Debiasing} 
\label{subsec:proposal}

We propose to define $g$ as a rescaled version of $f$ to better compare $g(X)$ with $f(X)$. Hence, the idea of our approach \emph{Ratio-Based Model Debiasing} (RBMD) is to edit the score $f(X)$ into a score $g(X)$, whose prediction $\hat{Y}^g$ is more fair, through a multiplicative factor $r(X)$:
\vspace{-0.1cm}
\begin{equation}
\label{eq: g(x)}
    g(X) = \sigma(r(X)f_{\text{logit}}(X)),
    \vspace{-0.1cm}
\end{equation}
where $\sigma(\cdot)$ is the sigmoid function to ensure $g(X)\in[0,1]$ and $f_{\text{logit}}$ is the logit value of the black-box score $f$. \\
We then propose to model the ratio $r_{w_g}$ as a Neural Network with parameters $w_g$ that takes as input $X$, and $g_{w_g}$ the corrected classifier. An ablation study on the impact of the architecture of $r_{w_g}$ has been done in Appendix \ref{sec:appendix architecture}. Using this formulation, satisfying the constraint [3] (similarity between $f$ and $g$) can easily be done by adding a penalization term $\mathcal{L}_{\text{ratio}}$ to control the behavior of $r_{w_g}$. Although several functions could be considered, we focus in the rest of the paper on ensuring that the score $f(X)$ is not modified unnecessarily, by defining $\mathcal{L}_{\text{ratio}}(r_{w_g}(X)) = ||r_{w_g}(X) - 1||_2^2$. A study on the effectiveness of this term can be found in Appendix \ref{sec:appendix hyperparams}. To ensure fairness (constraint [2]), we propose to leverage the technique proposed by~\citet{zhang2018mitigating}, which relies on a dynamic reconstruction of the sensitive attribute from the predictions using an adversarial model $h_{w_h}:\mathcal{Y}\rightarrow \mathcal{S}$ to estimate and mitigate bias. The higher the loss $\mathcal{L}_S(h_{w_h}(r_{w_g}(X)f_{\text{logit}}(X)), S)$ of this adversary is, the more fair the predictions $g_{w_g}(X)$ are.
Furthermore, this allows us to mitigate bias in a fairness-blind setting (i.e. without access to the sensitive attribute at test time).
Thus, Equation~\ref{eq:optimization-pb} for Demographic Parity~\footnote{\citet{zhang2018mitigating} describes how to address Equalized Odds with their method by modifying the adversary $h_{w_h}$ to take as input the true labels $Y$, on top of the predictions $g_{w_g}(X)$. By adopting the same approach, our method can also be easily adapted.} becomes:
\vspace{-0.1cm}
\begin{equation}
    \begin{aligned}
        \min_{w_g}  \quad \mathbbm{E} [\mathcal{L}_Y(g_{w_g}(X), Y)] \quad \textrm{[1]}\\
        \textrm{s.t} \quad \mathbbm{E}[\mathcal{L}_S(h_{w_h}(r_{w_g}(X)f_{\text{logit}}(X)), S)] \geq \epsilon' \quad \textrm{[2]}\\
        \textrm{and} \quad \mathbbm{E}[\mathcal{L}_{\text{ratio}}(r_{w_g}(X))] \leq \eta' \quad \textrm{[3]}
    \end{aligned}
    \label{eq:optimisation-r}
    \vspace{-0.1cm}
\end{equation}
with $\epsilon', \ \eta' > 0$. In practice, we relax the problem~\ref{eq:optimisation-r} and we optimize the following:  
\vspace{-0.1cm}
\begin{equation}
\label{eq: loss training}
\begin{aligned}
    \argmin_{w_g}\max_{{w_h}} \  &\frac{1}{N} \sum_{i=1}^N \mathcal{L}_{Y}(g_{w_g}(x_i), y_i) \\
    &- \lambda_{\text{fair}}\mathcal{L}_{S}(h_{w_h}(r_{w_g}(x_i)f_{\text{logit}}(x_i)), s_i) \\
    &+ \lambda_{\text{ratio}}\mathcal{L}_{\text{ratio}}(r_{w_g}(x_i)),
\end{aligned}
\vspace{-0.1cm}
\end{equation}
with $\lambda_{\text{fair}}$ and $\lambda_{\text{ratio}}$ two hyperparameters.

\section{Experiments}
\label{sec:experiments}

After describing our experimental setting in Section~\ref{sec:protocol}, we propose various experiments to showcase the efficacy of our method along several dimensions: in Section~\ref{sec:pareto}, we evaluate our method in terms of accuracy-fairness trade-off; in Section~\ref{sec:n_changes}, we evaluate the number of changes performed to the predictions of $f$ for given levels of accuracy and fairness; finally, we conduct additional analyses in Section~\ref{sec:interpretability} on the interpretability of our approach.

\subsection{Experimental Setting}
\label{sec:protocol}

We use two commonly used datasets in the fairness literature \cite{hort2023bias}: Law School~\cite{wightman1998lsac} and COMPAS~\cite{angwin2016machine}. The sensitive attribute for both datasets is \emph{race}.
Most existing post-processing methods do not focus on the same model-agnostic and fairness-blind setting and are therefore not directly comparable. To prove the efficiency of our method, we therefore consider the following approaches:
\vspace{-0.05cm}

\textbf{LPP}~\cite{xian2024optimal}\textbf{.} As shown in Table~\ref{tab:comparison}, LPP (Linear Post-Processing) is a post-processing method that is sensitive unaware at test time and is model agnostic\footnote{LPP Implementation: \href{https://github.com/rxian/fair-classification }{fair-classification.}}.\vspace{-0.05cm}

\textbf{Oracle}~\cite{xian2024optimal}\textbf{.} In the same work, the authors of LPP propose a variation of their method in the fairness-aware setting. We refer to it as "Oracle", as it uses the sensitive at test time and to the best of our knowledge it is the state-of-the-art between the algorithms that use the sensitive at test time.
\vspace{-0.05cm}

\textbf{ROC}~\cite{ROC}\textbf{.} ROC (Reject-Option Classifier) is a post-processing method that adopts a label-flipping heuristic based on model confidence. ROC also requires access to the sensitive at test time.
\vspace{-0.05cm}

\textbf{AdvDebias}~\cite{zhang2018mitigating}\textbf{.} As discussed in Sec.~\ref{sec:ratio}, we use $AdvDebias$, an in-processing method, for additional comparison with our method. We use the implementation provided in the AIF360 library~\cite{aif360-oct-2018}.
\vspace{-0.05cm}


After splitting each dataset in $\mathcal{D}_{\text{train}}$ ($70\%$) and $\mathcal{D}_{\text{test}}$ ($30\%$), we use a Logistic Regression (trained on $\mathcal{D}_{\text{train}}$) as our black-box classifier. We then train RBMD and the aforementioned competitors on the same $\mathcal{D}_{\text{train}}$ while varying the values of the hyperparameters of each method to achieve various levels of fairness and accuracy scores, calculated over $\mathcal{D}_{\text{test}}$.


\begin{figure}[t]
\begin{minipage}[t]{0.475\columnwidth}
  \includegraphics[width=\linewidth]{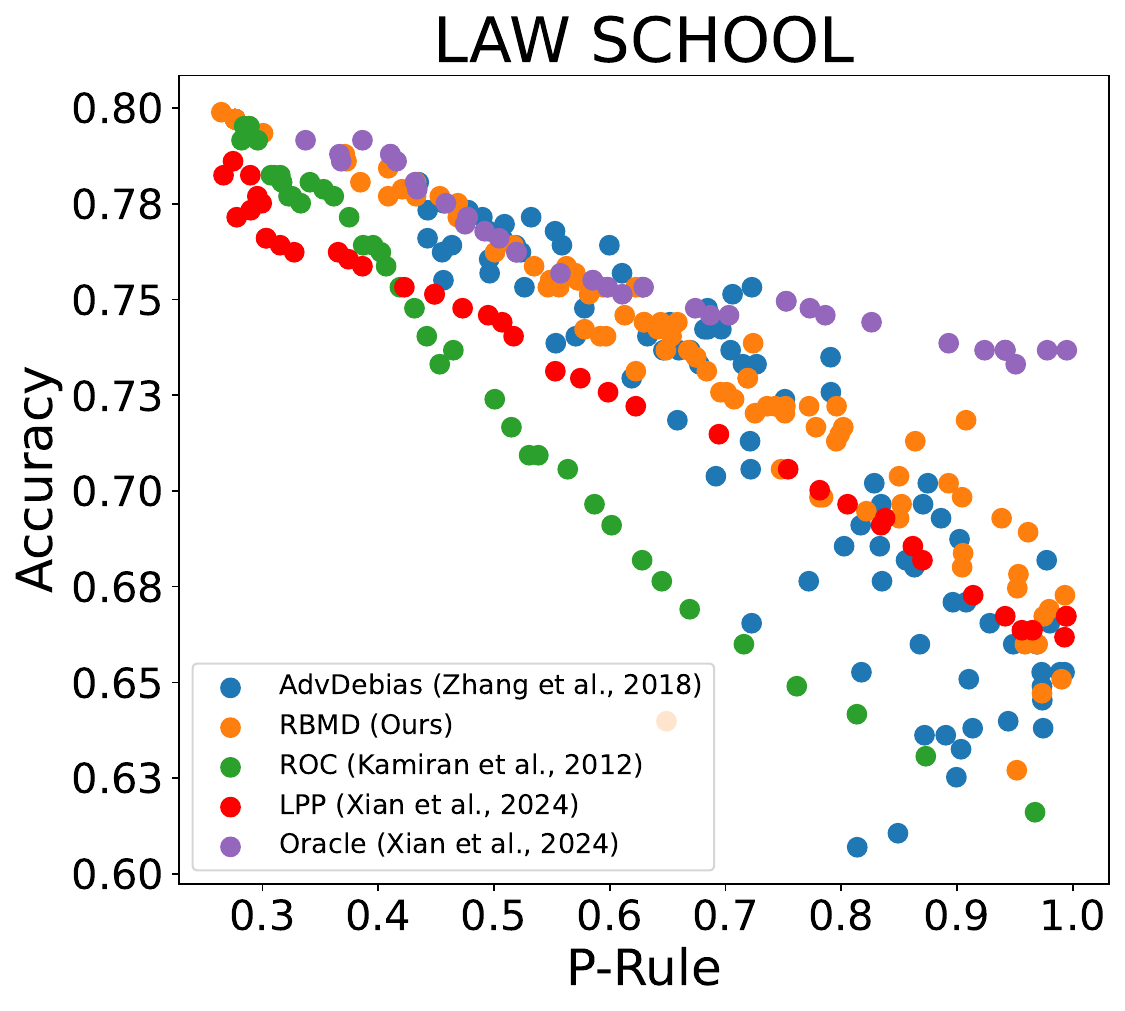}
\end{minipage}\hfill 
\begin{minipage}[t]{0.475\columnwidth}
  \includegraphics[width=\linewidth]{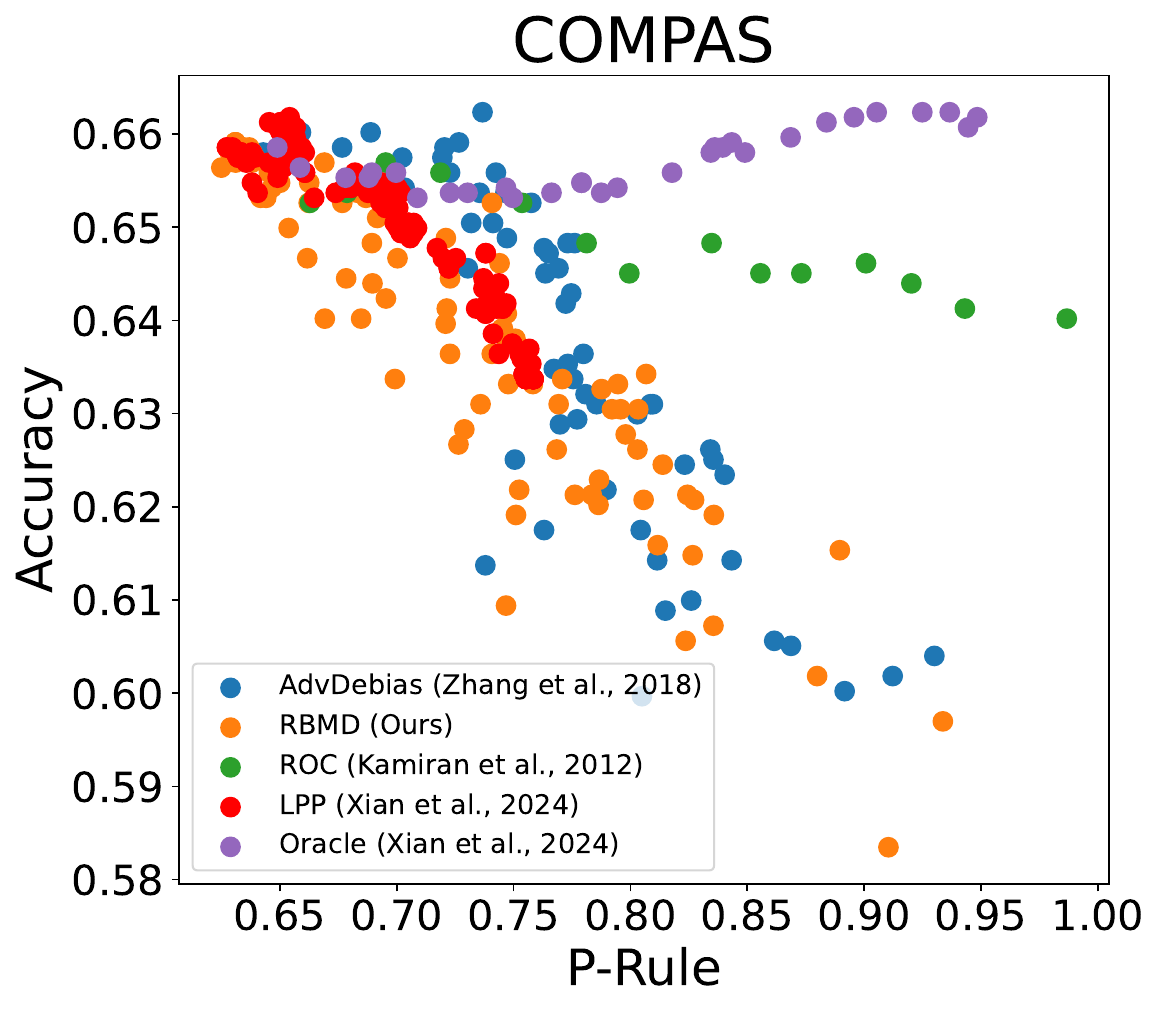}
\end{minipage}
\caption{Fairness vs Accuracy trade-off on Law School (left) and COMPAS (right) datasets.}
\label{fig:pareto}
\vspace{-0.5cm}
\end{figure}

\subsection{Results}
\label{sec:results}

\subsubsection{Experiment 1: Accuracy and Fairness}
\label{sec:pareto}

Figure~\ref{fig:pareto} shows the accuracy and fairness scores achieved by all competitors on the Law School and COMPAS datasets. We conduct each experiment three times for each parameter value to ensure robust results. From Figure~\ref{fig:pareto}, in which one dot corresponds to one model run, we can observe that our method outperforms LPP (that is model agnostic and that does not require sensitive at inference) on the Law School dataset and achieves the same results of it in the COMPAS dataset. Further, we show here that even by adding the constraint [3] in Equation \ref{eq:optimisation-r} RBMD can achieve almost similar results on both datasets as \textit{AdvDebias}, that has only reformulation of [1] and [2] in its optimization problem. Finally,  RBMD is outperformed by \textit{Oracle} (and also by ROC in COMPAS). However, this could be expected as these methods, on the contrary of RBMD and LPP, use the sensitive attribute at test time. Hence, these results show that RBMD is a competitive algorithm in terms of fairness-accuracy trade-off, regardless of its other added benefits that we describe in the next sections.

\subsubsection{Experiment 2: Number of changes}
\label{sec:n_changes}

\begin{figure}[t]
\centering
\vspace{-0.1cm}
  \includegraphics[width=0.95\columnwidth]{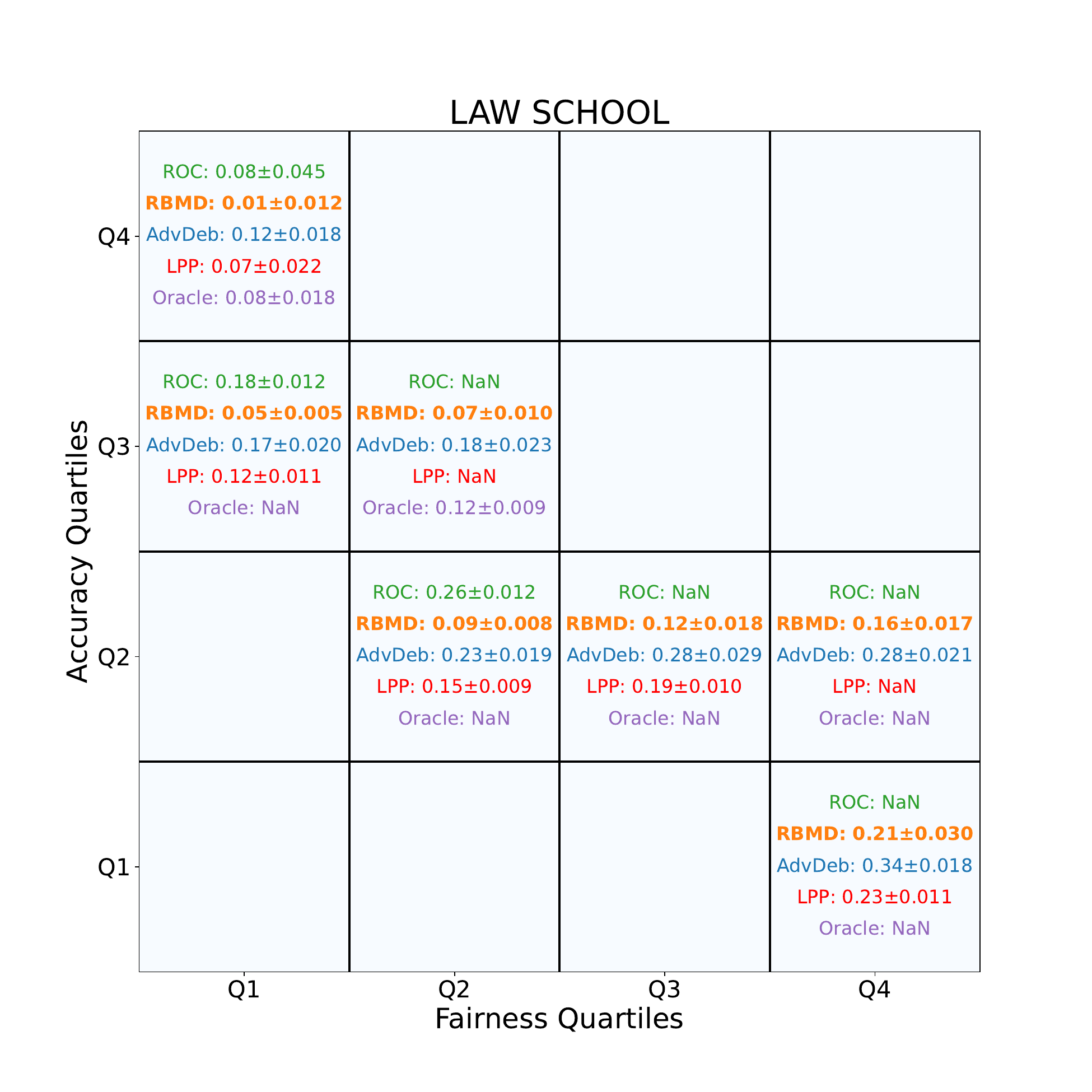}
\caption{Mean $\mathcal{P}$ ($\pm \text{std}$) performed by all methods. "NaN" means that no model falls into this range of scores. Only the cells where RBMD had at least 2 model runs are kept.}
\label{fig:n_changes}
\end{figure}

In this experiment, we measure the proportion $\mathcal{P}$ of changes between the predictions of the black-box model and the ones of the fairer model on the test set $\mathcal{D}_{\text{test}}$ . As the values $\mathcal{P}$ can take depend on the levels of fairness and accuracy of the model, we propose to measure $\mathcal{P}$ for models with \emph{comparable} levels of fairness and accuracy. To do so, we discretize the accuracy and fairness scores into 4 segments each, defined as quartiles of the values reached by $AdvDebias$ (see Appendix~\ref{sec:appendix quartiles} for details). This defines a $4 \times 4$ grid, describing a discretized version of the Pareto plot shown in Figure~\ref{fig:pareto}. 
We show in Figure~\ref{fig:n_changes} the average values of $\mathcal{P}$ obtained for the runs of the considered methods falling into each cell of this grid. We observe that regardless of the fairness and accuracy levels considered, RBMD consistently achieves lower values of changes $\mathcal{P}$ compared to its competitors, both for Law School (Figure \ref{fig:n_changes}) and for COMPAS (Figure \ref{fig:n_changes compas}) dataset.



\subsubsection{Experiment 3: Explaining prediction changes}
\label{sec:interpretability}

\begin{figure}[t]
\centering
  \includegraphics[width=0.6\columnwidth]{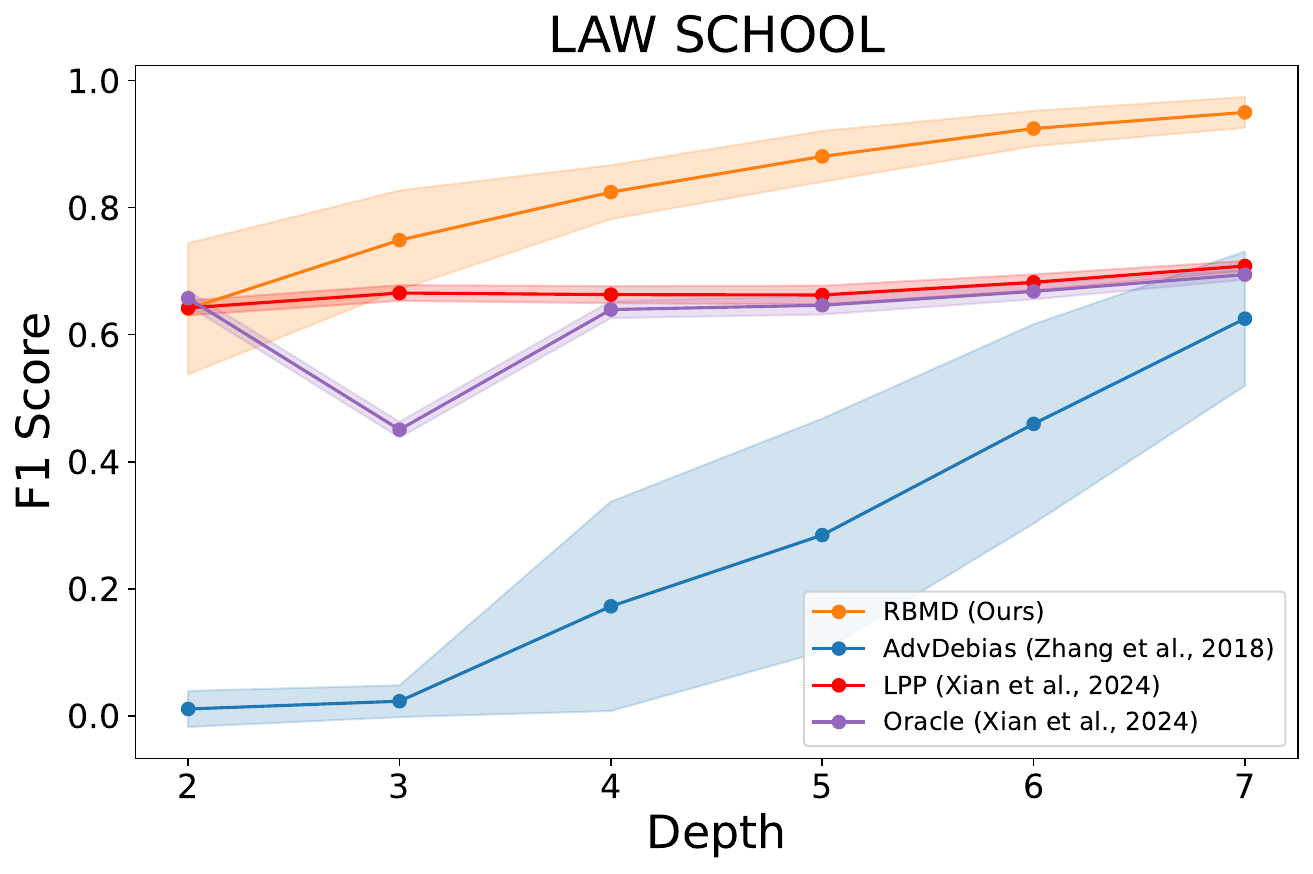}
\caption{\textit{F1-score} of CART for different depths for similar values of fairness (Q3) and accuracy (Q2). There is no point for ROC in this cell, so it is not included in the plot.}
\label{fig: Locality}
\end{figure}

In this experiment, we aim to show how RBMD facilitates understanding \emph{which} instances were targeted by the debiasing models.
For this purpose, given a model $g$, we train a surrogate decision tree model on the dataset $\mathcal{D}_{\text{train}}$ with labels $\mathbbm{1}_{\hat{Y}_{\text{train}}^f \neq\hat{Y}_{\text{train}}^g}$. Such a decision tree (e.g. the one in Figure~\ref{fig:DT} in Appendix), defines segments of instances targeted by a bias mitigation method (here RBMD) in an interpretable manner. In Figure~\ref{fig: Locality} is shown the F1 score achieved by all decision trees trained for debiasing models with comparable performances, depending on their depth. A high F1 score means that the decision tree accurately describes the instances with prediction changes. 
We observe that the instances targeted by RBMD can generally be described with simpler decision trees, suggesting that they lie in a lower number of different regions of the feature space. This opens the possibility of better interpreting \textit{which} are the instances targeted by the debiasing algorithm. Another experiment in this direction is presented in Appendix \ref{sec:appendix architecture}.



\section{Conclusion}
\label{sec:conclusion}
In this ongoing work, we presented RBMD, an innovative post-processing technique designed to adjust the predictions of biased models, ensuring fairness with minimal changes to the original predictions and without significant accuracy loss compared to other methods. Beyond further experimental validation, future works include adapting the proposed approach to achieve Equalized Odds.


\bibliography{example_paper}
\bibliographystyle{icml2024}

\newpage
\appendix
\onecolumn

\section{Implementation details}
\label{sec:appendix implementatio details}

\subsection{Architecture of $r$ and Interpretability}
\label{sec:appendix architecture}
In this subsection, we do an ablation study on the architecture of the ratio network $r_{w_g}(\cdot)$ on the Law School dataset. For the experiments and the plots shown in the paper, the architecture of the network $r_{w_g}(\cdot)$ consists of a fully connected neural network with two hidden layers with a \textit{ReLU} activation functions after each linear. The final layer outputs a single value that is the ratio to be multiplied with $f_{\text{logit}}(X)$.\\
In order to look at the impact of such an architecture on the results of Figure \ref{fig:pareto}, we train our RBMD method with architectures of $r_{w_g}(\cdot)$ with different power. Specifically, we choose to increase the hidden layers from two to three and to decrease them to zero. In this last setting, we simply have a linear ratio
\begin{equation}
\label{eq:linear ratio}
    r_{w_g}(X) = w_0 + w_1 x_1 + ... + w_d x_d.
\end{equation}
Hence, we would be able to leverage more the interpretability of our debiasing method. Notice that we carried out these experiments on the \textit{Law School} dataset while keeping fixed the train-test split of the dataset.

\begin{figure}[h]
\begin{minipage}[t]{0.475\columnwidth}
  \centering
  \includegraphics[width=\linewidth]{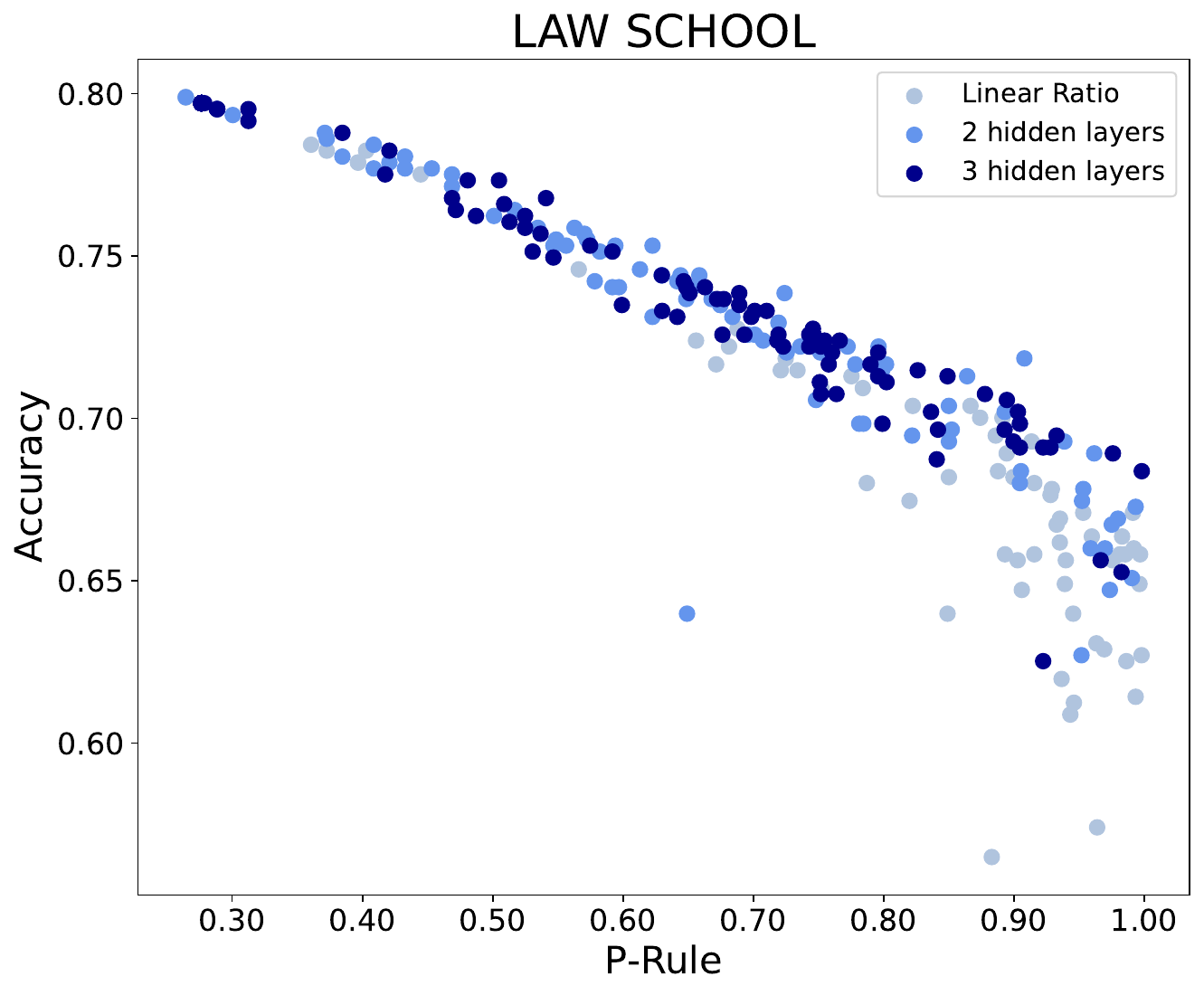}
  \caption{Fairness vs Accuracy trade-off on Law School for different ratio architectures. Each dot corresponds to one model run.}
  \label{fig: pareto architecture}
\end{minipage}\hfill 
\begin{minipage}[t]{0.475\columnwidth}
  \centering
  \includegraphics[width=\linewidth]{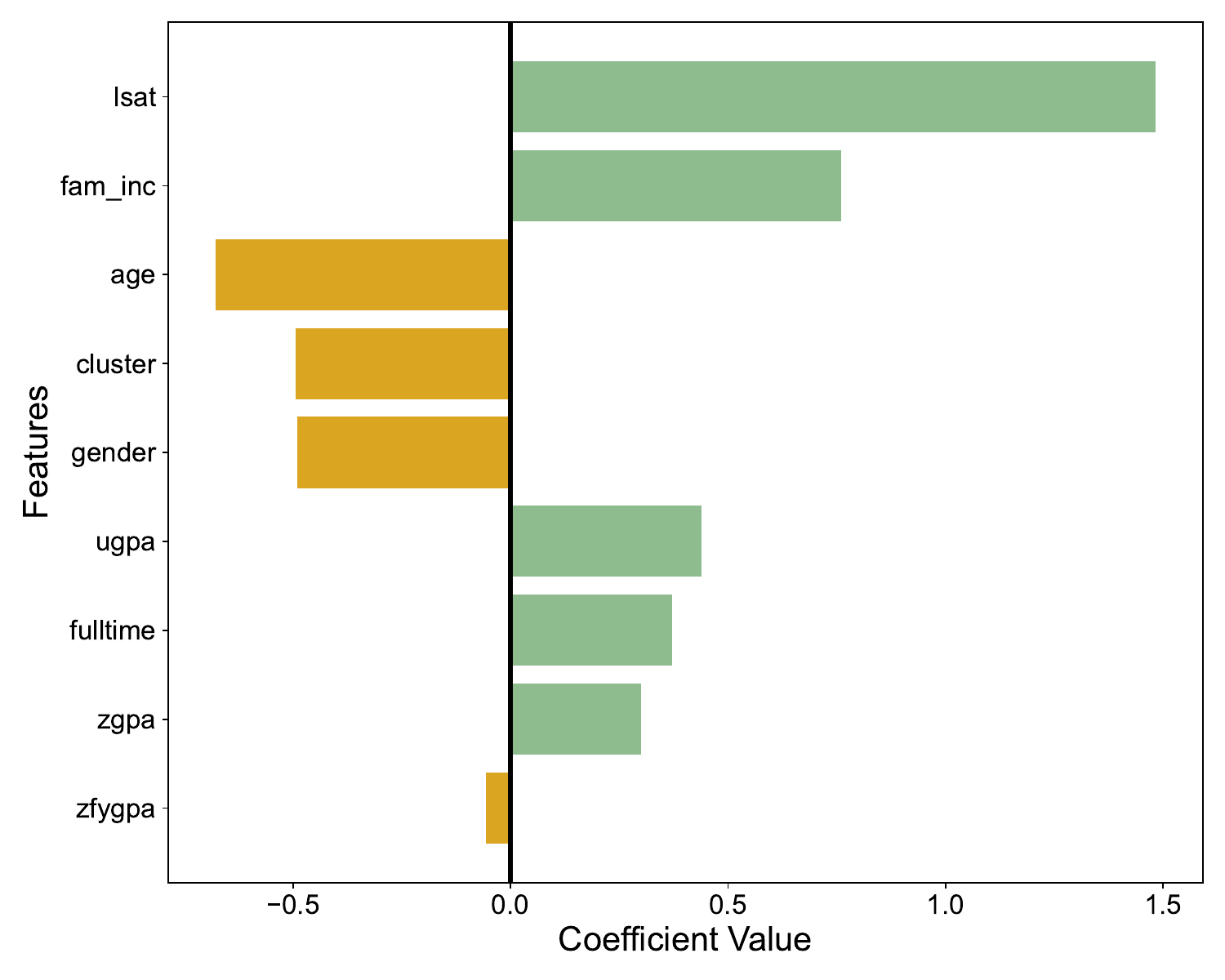}
  \caption{Weights $w_1,...,w_d$ of Equation \ref{eq:linear ratio} when the ratio architecture is linear.}
  \label{fig: barchart weights}
\end{minipage}
\end{figure}

From Figure \ref{fig: pareto architecture} we can observe that the performances of the RBMD method in the Fairness vs Accuracy plot do not change much by changing the architecture of the ratio network. For this reason and the reasons stated above, we decided to dive deeper into the linear ratio. \\
Hence, we select an experiment for which the debiased model has $(\text{fairness}, \ \text{accuracy})=(0.78, \ 0.70)$, obtained with $\lambda_{\text{ratio}} = 0.25$ and $\lambda_{\text{fair}} = 3$. For such a model, we then plot the learned weights of the network, that corresponds to $w_0, ..., w_d$ in Equation \ref{eq:linear ratio}, where $\beta_0$ refers to the weights of the intercept.\\
To better use the plot in \ref{fig: barchart weights} in support of the interpretation of the debiasing algorithm, it is important to look at the values that each feature can take since you can notice from Equation \ref{eq: g(x)} that only a negative ratio will swap a prediction. Hence, a feature that is pushing towards a switch in the prediction is a feature whose coefficient in Figure \ref{fig: barchart weights} is negative because the standardization we do in the data has the effect of making every feature's distribution with only positive support.

\subsection{Calibration}

\begin{figure}[t]
\centering
  \includegraphics[width=0.45\linewidth]{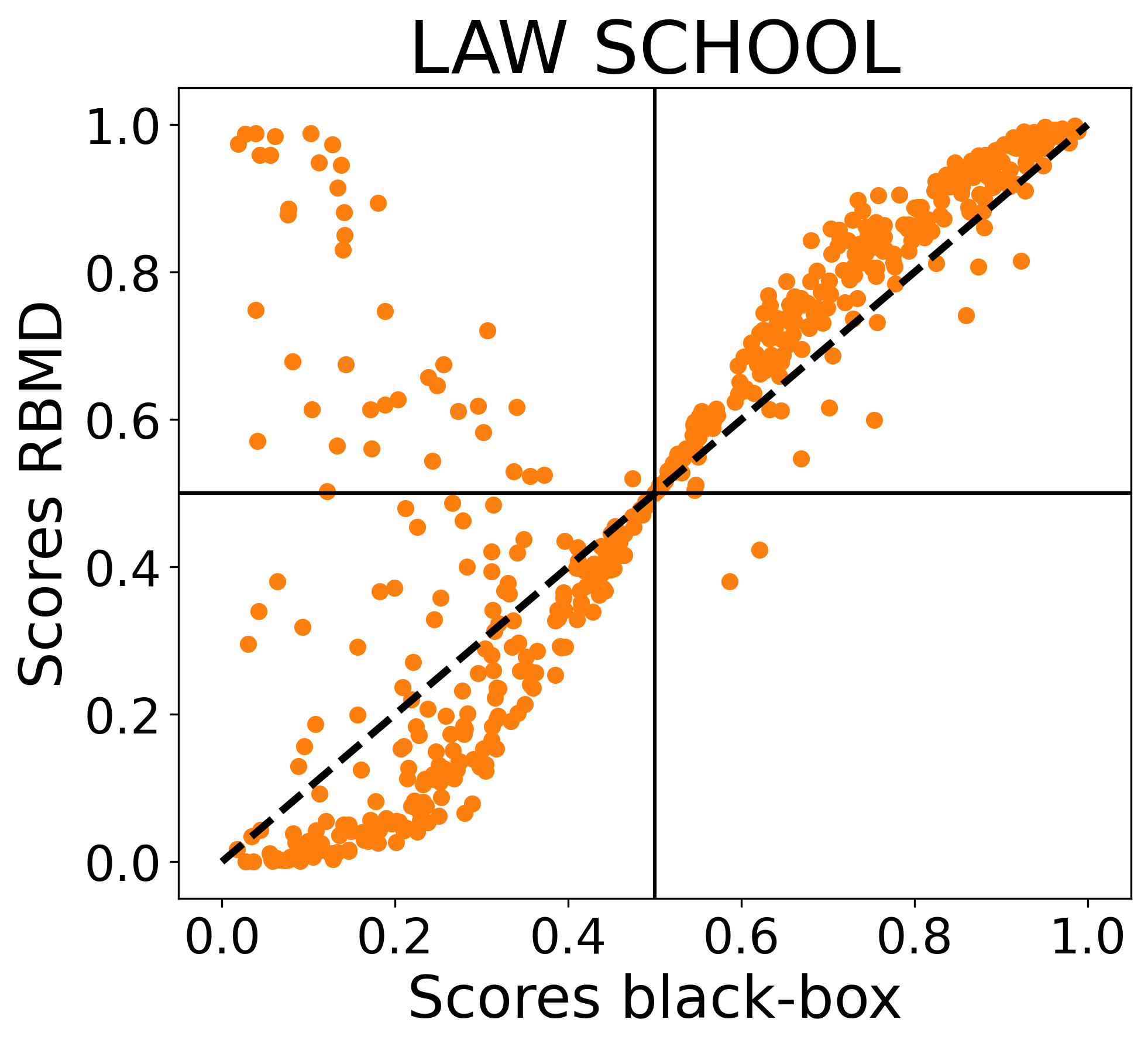}
  \includegraphics[width=0.45\linewidth]{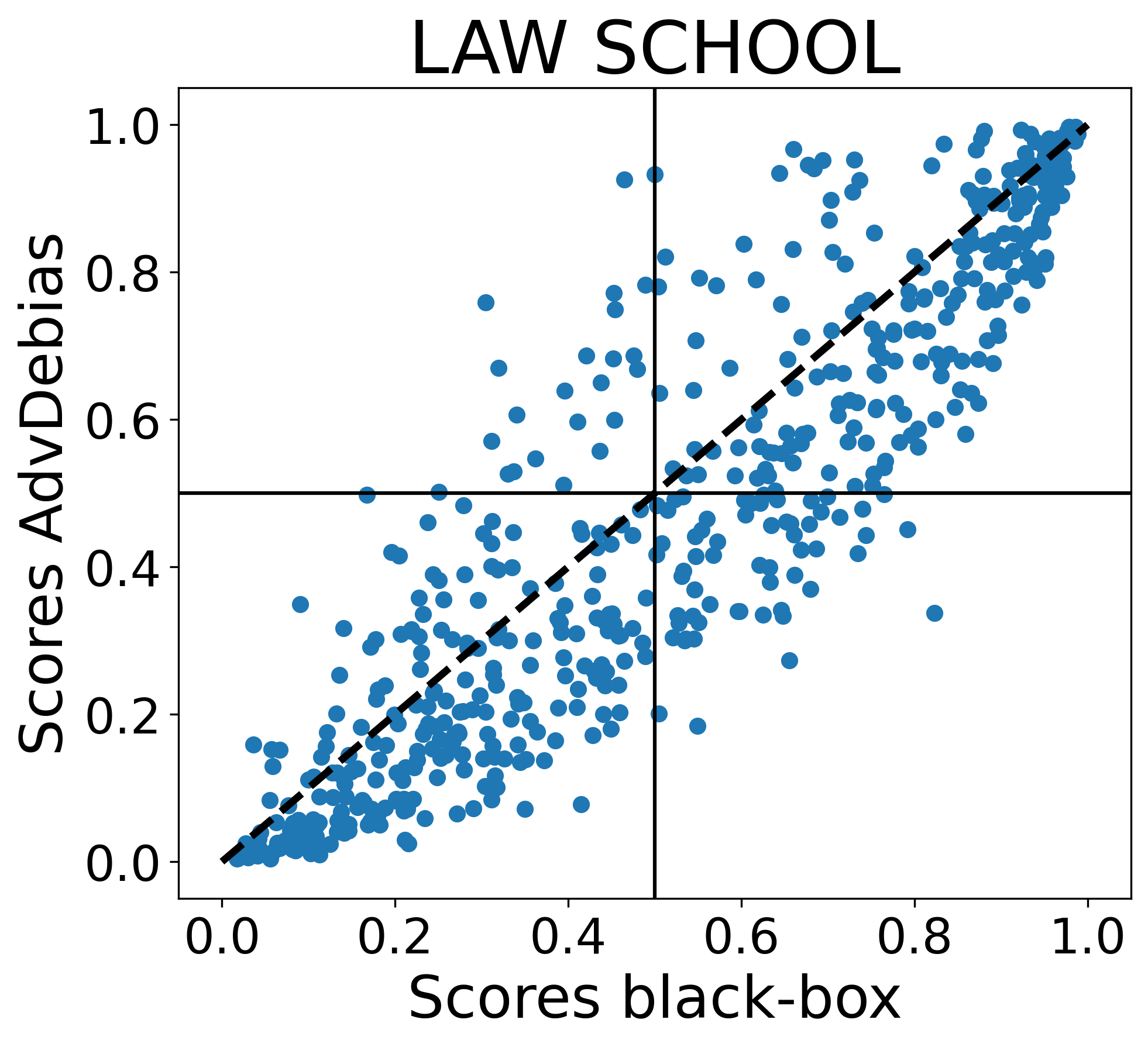}
\caption{Calibration plots of RBMD (left) and AdvDebias (right) for similar values of (fairness, accuracy): AdvDebias (0.65, 0.74), RBDM (0.68, 0.73).}
\label{fig:calibration plot}
\end{figure}

To give further insights on the differences between models after debiasing, we show in Figure~\ref{fig:calibration plot} the distribution of the classification probabilities returned by the blackbox model (x-axis) and a fairer one (y-axis) over the test set, for RMBD and $AdvDebias$ for reference. We directly observe the impact of the MSE constraint in $\mathcal{L}_{\text{ratio}}$: most of the classification probabilities $g(x)$ are kept similar to their initial value $f(x)$, i.e. along the diagonal. Furthermore, this gives insights into the strategy adopted by the debiasing method: we thus observe that to achieve fairness, the main changes performed by RBMD are positive changes, i.e. from class $0$ to class $1$ (top left quadrant).

\subsection{Loss and Hyperparameters}
\label{sec:appendix hyperparams}
As with every deep learning-based method, it is crucial to play with hyperparameters $\lambda_{\text{fair}} \text{ and } \lambda_{\text{ratio}}$ to adapt the weights of the terms in the loss function we use for our Ratio-Based Model Debiasing. For this reason, we evaluate the performance of our method over a grid search of combinations of parameters $(\lambda_{\text{fair}}, \ \lambda_{\text{ratio}})$.

\paragraph{Effectiveness of $\mathcal{L}_{\text{ratio}}$.} As a first step, we validate the effectiveness of $\lambda_{\text{ratio}}\mathcal{L}_{\text{ratio}}(r_w(x_i))$ in Equation \ref{eq: loss training}. Recall that $\mathcal{L}_{\text{ratio}}$ has been added to decrease the number of changes with respect to the biased model.
To do so, we choose the \textit{Law School} dataset and we fixed a value of $\lambda_{\text{fair}}$; we then vary the value of $\lambda_{\text{ratio}}$ while observing the distribution of the ratios on the test set $\mathcal{D}_{\text{test}}$ for all these different values of $\lambda_{\text{ratio}}$.
\begin{figure}[h]
\centering
  \includegraphics[width=0.7\columnwidth]{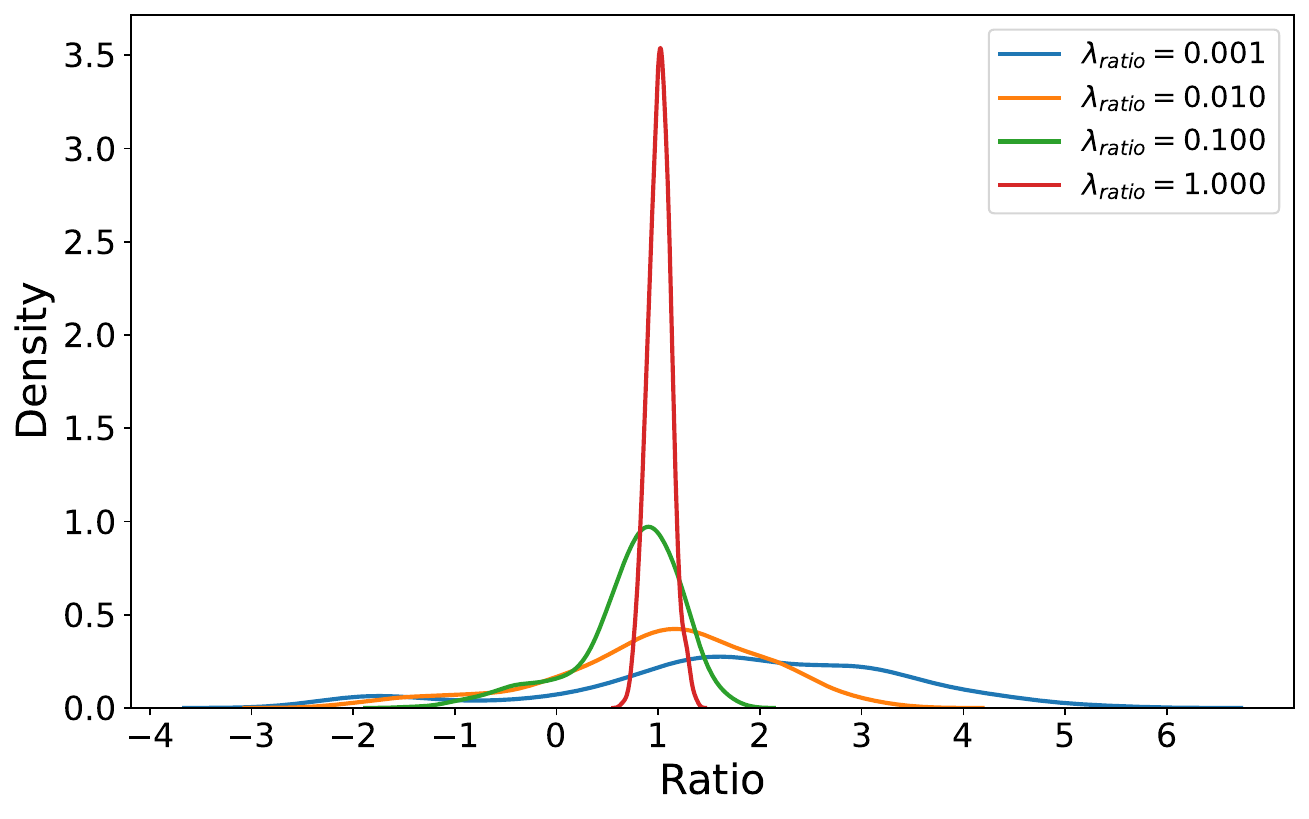}
\caption{Ratio distribution on the test set for different values of $\lambda_{\text{ratio}}$ while keeping fixed $\lambda_{\text{fair}} = 1.5$.}
\label{fig: ratio distr appendix}
\end{figure}
As we can observe from Figure \ref{fig: ratio distr appendix}, by increasing the value of $\lambda_{\text{ratio}}$ we make the distribution of the ratios on the test set more skewed around one, meaning that RBMD is less prone to change a prediction. If we then decrease $\lambda_{\text{ratio}}$, we increase the probability of a change in the prediction since the support of the probability distribution has negative numbers. We recall once again that the negativity of the ratio is necessary to have a switch in the prediction. Finally, the skewness around one of the ratios' distribution helps us not change all the scores of the biased model as you can also observe from Figure \ref{fig:calibration plot}. In such a plot, in fact, we can observe how the scores coming from RBMD are more aligned on the diagonal. Notice that a perfect alignment on the diagonal would mean that every score is unchanged after the debiasing. Finally, the plot in Figure \ref{fig:calibration plot} also shows in which direction the prediction is switched since the top-left and bottom-right quadrants are the ones where we have switched a prediction.

\paragraph{Hyperparameters value.} For the \textit{Law School} dataset we did a grid search over $\lambda_{\text{ratio}} \in [0, \ 0.25]$ and $\lambda_{\text{fair}} \in [0, \ 3]$. \\
For the \textit{COMPAS} dataset we did a grid search over $\lambda_{\text{ratio}} \in [0, \ 4]$ and $\lambda_{\text{fair}} \in [0, \ 0.1 ]$.

\section{Quartiles definition}
\label{sec:appendix quartiles}
In the paper, we referred several times to the quartiles of fairness and accuracy. We used them in order to compare the methods in the region of the Pareto for which they have approximately the same levels of fairness and accuracy. These quartiles have been computed on the \textit{AdvDebias} method and their values are shown in the tables below.
\subsection{Law School Dataset}
\begin{table}[h!]
    \centering
    \begin{tabular}{|c|c|}
        \hline
        Fairness Quartile & Value range \\
        \hline
        Q1 & [$0, \ 0.559$] \\
        Q2 & [$0.559, \ 0.721$] \\
        Q3 & [$0.721, \ 0.872$] \\
        Q4 & [$0.872, \ 1$] \\
        \hline
    \end{tabular}
    \caption{Fairness quartiles and their ranges of values (\textit{P-Rule}) for Law School dataset.}
    \label{tab:fairness_quartiles}
\end{table}

\begin{table}[h!]
    \centering
    \begin{tabular}{|c|c|}
        \hline
        Accuracy Quartile & Value range \\
        \hline
        Q1 & [$0, \ 0.671$] \\
        Q2 & [$0.671, \ 0.729$] \\
        Q3 & [$0.729, \ 0.755$] \\
        Q4 & [$0.755, \ 1$] \\
        \hline
    \end{tabular}
    \caption{Accuracy Quartiles and their ranges of values for Law School Dataset.}
    \label{tab:accuracy_quartiles}
\end{table}

\subsection{COMPAS Dataset}
\begin{table}[h!]
    \centering
    \begin{tabular}{|c|c|}
        \hline
        Fairness Quartile & Value range \\
        \hline
        Q1 & [$0, \ 0.735$] \\
        Q2 & [$0.735, \ 0.769$] \\
        Q3 & [$0.769, \ 0.806$] \\
        Q4 & [$0.806, \ 1$] \\
        \hline
    \end{tabular}
    \caption{Fairness quartiles and their ranges of values (\textit{P-Rule}) for COMPAS dataset.}
    \label{tab:fairness_quartiles}
\end{table}

\begin{table}[h!]
    \centering
    \begin{tabular}{|c|c|}
        \hline
        Accuracy Quartile & Value range \\
        \hline
        Q1 & [$0, \ 0.625$] \\
        Q2 & [$0.625, \ 0.639$] \\
        Q3 & [$0.639, \ 0.654$] \\
        Q4 & [$0.654, \ 1$] \\
        \hline
    \end{tabular}
    \caption{Accuracy Quartiles and their ranges of values for COMPAS dataset.}
    \label{tab:accuracy_quartiles}
\end{table}

\section{Further results}
\label{sec:appendix results}

\subsection{Law School dataset}
Following, you can find the CART decision tree trained as explained in Section \ref{sec:interpretability}.
\begin{figure}[h]
\centering
  \includegraphics[width=0.4\columnwidth]{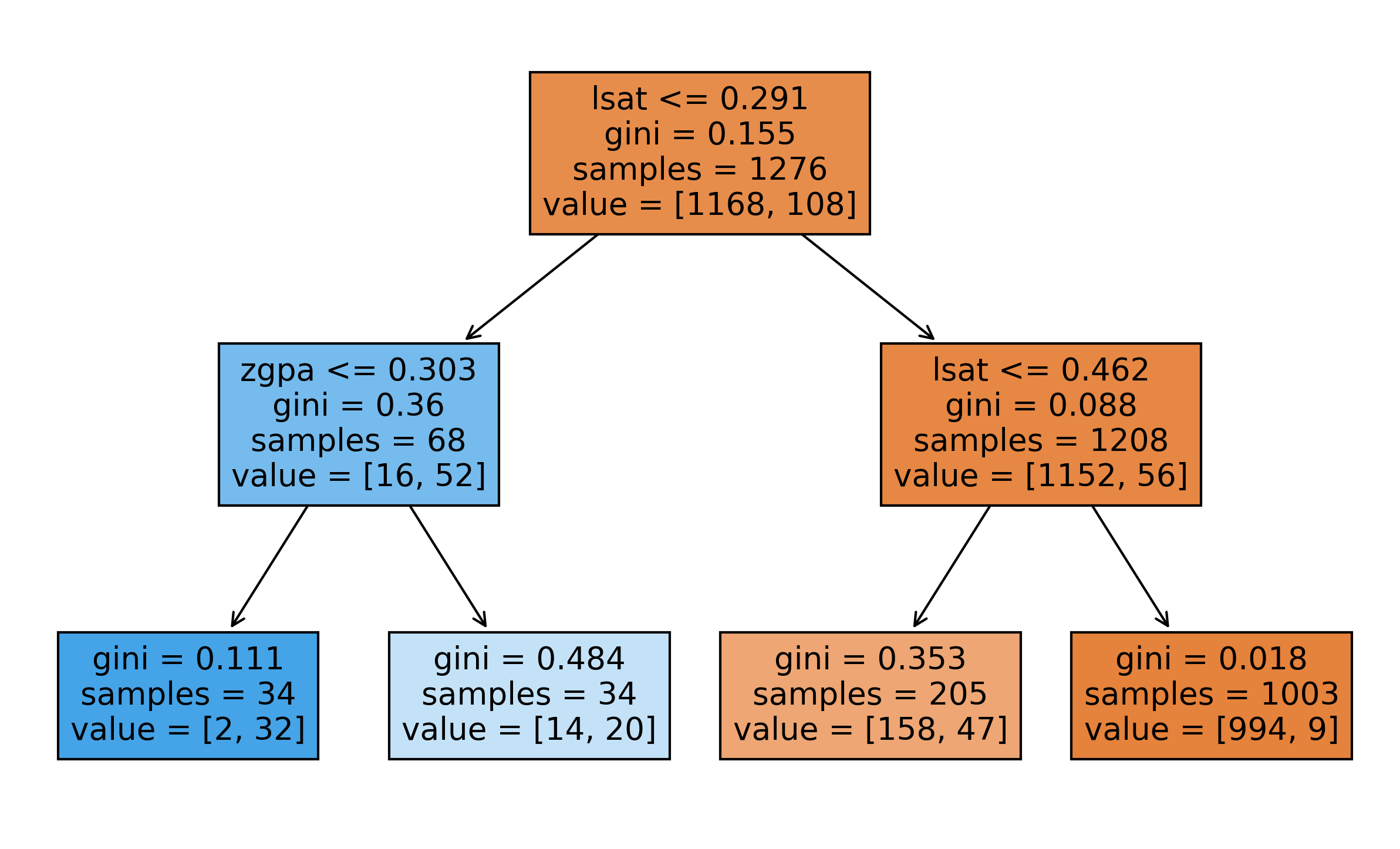}
\caption{Decision tree of depth $2$ for the RBMD algorithm on Law School dataset at a level of (fairness, accuracy) of (0.68, 0.73).}
\label{fig:DT}
\end{figure}
From this plot, we can observe that the sub-regions of the features space are identified by the variables \textit{cluster}, \textit{fam\_inc}, and \textit{lsat}. In particular, we can here observe for example the leaf $cluster\leq 0.9 \rightarrow lsat\leq  0.47$ for which we have a prediction equals to one, i.e. a difference between biased and unbiased prediction.

\subsection{COMPAS dataset}
Following, you can find the results of Section \ref{sec:n_changes} for COMPAS dataset. Again, we measure the proportion $\mathcal{P}$ of prediction changes on the test sets between the black-box model and the fairer ones for every method introduced in Section \ref{sec:experiments}.
\vspace{-0.3cm}
\begin{figure}[h]
\centering
  \includegraphics[width=0.4\columnwidth]{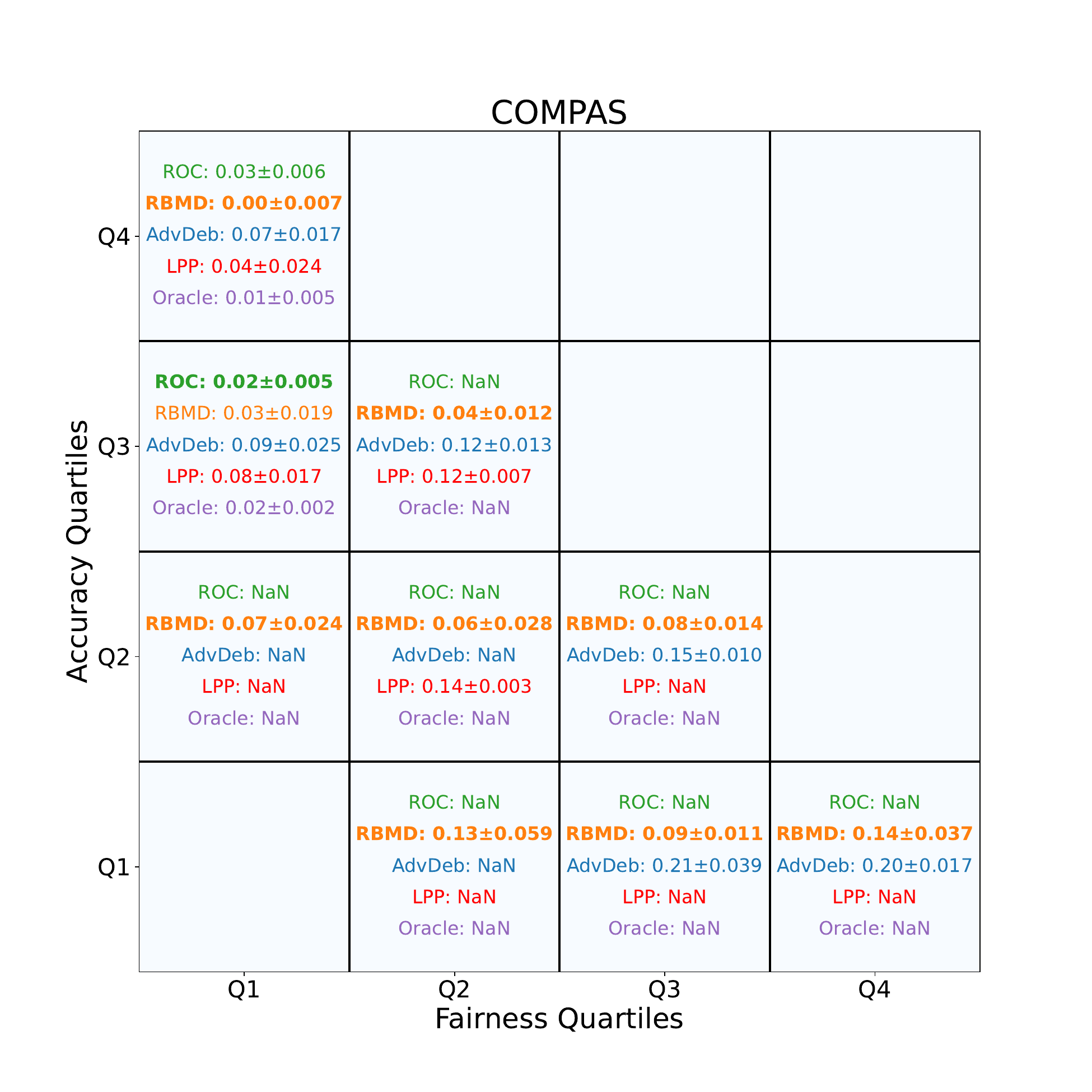}
\caption{Mean $\mathcal{P}$ ($\pm \text{std}$) performed by all methods. "NaN" means that no model falls into this range of scores. Only the cells where RBMD had at least 2 model runs are kept.}
\label{fig:n_changes compas}
\end{figure}

Figure \ref{fig:n_changes compas} shows how the (mean) percentage of changes $\mathcal{P}$ between biased and unbiased predictions varies over different levels of fairness and accuracy. As already shown in Figure \ref{fig:n_changes} for the Law School dataset, RBMD achieved the best results on this desideratum. For the COMPAS dataset, there is one cell (Q1 fairness vs Q3 accuracy) for which \textit{Oracle} has lower (mean) $\mathcal{P}$ but this difference is not statistically significant due to the high value of the standard deviation of the $\mathcal{P}$ of RBMD in that cell.


\end{document}